\documentclass{article}
\pdfoutput=1

\usepackage{arxiv}

\usepackage{amsmath,amsfonts,bm}

\def\eqref#1{equation~\ref{#1}}

\def\1{\bm{1}}

\def\vh{{\bm{h}}}

\def\vx{{\bm{x}}}

\def\mH{{\bm{H}}}

\def\mP{{\bm{P}}}

\def\mS{{\bm{S}}}

\def\mW{{\bm{W}}}

\def\mZ{{\bm{Z}}}

\DeclareMathAlphabet{\mathsfit}{\encodingdefault}{\sfdefault}{m}{sl}
\SetMathAlphabet{\mathsfit}{bold}{\encodingdefault}{\sfdefault}{bx}{n}
\newcommand{\tens}[1]{\bm{\mathsfit{#1}}}

\def\tH{{\tens{H}}}

\def\tW{{\tens{W}}}

\def\sR{{\mathbb{R}}}

\usepackage[utf8]{inputenc} 
\usepackage[T1]{fontenc}    
\usepackage{hyperref}       
\usepackage{url}            
\usepackage{booktabs}       
\usepackage{amsfonts}       
\usepackage{nicefrac}       
\usepackage{microtype}      
\usepackage{xcolor} 
\usepackage{amsmath} 
\usepackage{graphicx} 
\usepackage{bbm} 
\usepackage{censor} 
\usepackage{subfig}
\usepackage{enumitem}
\usepackage{booktabs}
\usepackage[backend=bibtex,style=authoryear, maxcitenames=1]{biblatex}
\let\tb\textbf 

\title{Concentric Spherical GNN for 3D \\ Representation Learning}

\author{%
  James Fox\thanks{Corresponding author. A significant part of this work was completed during internship at Sandia National Laboratories.} \\
  Georgia Institute of Technology\\
  \texttt{jfox43@gatech.edu} \\
  \And
  Bo Zhao \\
  Georgia Institute of Technology \\
  \texttt{bzhao68@gatech.edu} \\
  \AND
  Sivasankaran Rajamanickam \\
  Sandia National Laboratories \\
  \texttt{srajama@sandia.gov} \\
  \And
  Rampi Ramprasad \\
  Georgia Institute of Technology \\
  \texttt{rampi.ramprasad@mse.gatech.edu} \\
  \And
  Le Song \\
  Mohamed bin Zayed University \\ of Artificial Intelligence \\
  \texttt{le.song@mbzuai.ac.ae} \\
}

\let\tb\textbf 

\begin{document}

\maketitle

\begin{abstract}
Learning 3D representations that generalize well to arbitrarily oriented inputs is a challenge of practical importance in applications varying from computer vision to physics and chemistry.
We propose a novel multi-resolution convolutional architecture for learning over concentric spherical feature maps, of which the single sphere representation is a special case.
Our hierarchical architecture is based on alternatively learning to incorporate both intra-sphere and inter-sphere information. 
We show the applicability of our method for two different types of 3D inputs, mesh objects, which can be regularly sampled, and point clouds, which are irregularly distributed. 
We also propose an efficient mapping of point clouds to concentric spherical images, thereby bridging spherical convolutions on grids with general point clouds.
We demonstrate the effectiveness of our approach in improving state-of-the-art performance on 3D classification tasks with rotated data.
\end{abstract}

\section{Introduction}
\label{sec:intro}







While convolutional neural networks have been applied to great success to 2D images, extending the same success to geometries in 3D has proven more challenging. 
A desirable property and challenge in this setting is to learn descriptive representations that are also equivariant to any 3D rotation. 
\textcite{cohen2018spherical} and  \textcite{esteves2018learning} showed that the spherical domain permits rotationally equivariant convolutions, resulting in the Spherical Convolutional Neural Network (CNN).
Subsequent works from \textcite{jiang2018spherical}, \textcite{cohen2019gauge}, \textcite{defferrard2020deepsphere} improved on the scalabilty of these convolutions on more uniformly sampled spherical grids.

Existing Spherical CNNs operate over feature maps resulting from projection and sampling of data onto the sphere.
We show that it is more expressive and general to instead operate over multiple, concentric spheres to represent 3D data.
Our main innovation is introducing a new two-phase convolutional scheme for learning over a concentric spheres representation, by alternating between inter-sphere and intra-sphere convolutions. 
We use graph convolutions to incorporate intra-sphere information and propose radial convolutions to incorporate inter-sphere information. 
Our proposed architecture is hierarchical, based on the highly uniform icosahedron discretization of the sphere.
Combining intra-sphere and inter-sphere convolutions has a conceptual analogy to gradually incorporating information volumetrically.
At the same time, our choice of convolutions allows retaining the rotational equivariance properties of Spherical CNNs.

We demonstrate the effectiveness and generality of our approach through two 3D classification experiments with different types of input data: mesh objects and general point clouds.
The latter poses an additional challenge for discretization-based methods, as native point clouds are non-uniformly distributed in 3D space.
We improve state-of-the-art performance in the challenging and most general scenario of training and testing on arbitrarily rotated data.

To summarize our contributions: 
\begin{enumerate}[nolistsep,nosep]
\item
We propose a new multi-sphere icosahedral discretization for representation of 3D data, and show that incorporating the radial dimension can greatly enhance representation ability over single-sphere representations.

\item
We also introduce a novel convolutional architecture for multi-sphere discretization by introducing two different types of convolutions, graph convolution and radial convolutions, for incorporating intra-sphere and inter-sphere information.
Their combined use leads to an expressive architecture that is also rotationally equivariant.
Our proposed convolutions are also scalable, being linear with respect to the discretization size.

\item  
We design mappings of both 3D mesh objects and general point clouds to the proposed representation.
We achieve new state-of-the-art performance on ModelNet40 point cloud classification, using the proposed model and an input data mapping based on radial basis functions.
We also outperform existing Spherical CNN performance in SHREC17 3D mesh classification.
\end{enumerate}

\section{Related Work}
{\bf Spherical CNNs.} 
Spherical CNNs were introduced in \textcite{cohen2018spherical} and \textcite{esteves2018learning} for learning rotationally equivariant representations of data defined on the sphere.
The key contribution was the definition of spherical convolutions that permit rotational equivariance to general 3D rotations.
Subsequent works have improved on the complexity of the convolutions in conjunction with the type of spherical discretization considered. 
\textcite{jiang2018spherical} proposed parameterized differential operators, which achieve linear complexity with respect to grid resolution but equvariance is restricted to $z$-axis aligned rotations.
\textcite{cohen2019gauge} and \textcite{defferrard2020deepsphere} proposed gauge equivariant convolutions and graph convolutions respectively, which also have linear complexity and are equivariant to general rotations.
These prior works rely on sampling and/or projection of data at input to a single sphere, which we argue is too restrictive for data representations such as arbitrary point clouds.

Some other works have additionally attempted to improve expressiveness of Spherical CNNs for point cloud applications.
\textcite{rao2019spherical} uses an additional parameterized module that learns to first map groups of points to features on vertices prior to spherical convolutions.
Instead of learning an initial mapping to a single sphere, which is somewhat orthogonal in concept to our main contribution, we define convolutions over concentric spheres that enable better capturing relationships between points through all stages of the network.
\textcite{you2018prin} proposed extending convolutions to a spherical voxel grid, explicitly incorporating an additional radial dimension.
However their method is specific to higher complexity SO3 convolutions on a non-uniform spherical grid.
Our work combines the benefits of linear complexity convolutions on uniform spherical discretization with learning more expressive representations via distinct intra-sphere and inter-sphere convolutions, and obtain much better results in practice.

{\bf Pointwise Convolution Networks.}
Unlike Spherical CNNs, which operate on a discretization, point-convolution networks operate directly on raw points. Early work such as \textcite{qi2017pointnet} proposed learning permutation invariant functions that use point coordinates as initial features. 
While descriptive, coordinates are not rotationally equivariant and such architectures do not generalize well to arbitrary rotations.
\textcite{chen2019clusternet}, \textcite{sun2019srinet}, \textcite{zhang2019rotation} address the problem by substituting point coordinates with rotationally invariant features, by carefully using points for frames of reference and then extracting information such as distances and angles.
These are then input into backbone convolutional architectures for point-wise convolutions.
On the other hand, 
\textcite{thomas2018tensor} and \textcite{poulenard2019effective} propose pointwise convolutional filters based on spherical harmonic functions to directly learn from points in a rotational equivariant manner.
Point-wise convolution methods scale linearly with respect to data points by restricting to fixed size neighborhoods (e.g. via $K$-nearest neighbors). 
While generalizing well to unseen rotations, overall these methods perform significantly worse than their counterparts (including our work) on less challenging rotation settings.
We surmise this could be due to difficulty in determining an expressive and robust convolutional hierarchy with respect to irregularly distributed data points to offset the loss of coordinate information, compared to a refinable regular discretization such as the one used in our work. 


\section{Desiderata for 3D Representation Learning}
Two basic desiderata of a model for 3D data are (1) expressiveness of learned representation and (2) robustness to arbitrary orientations/rotations of the data.
To summarize, Spherical CNNs provide a solution to the latter by introducing rotationally equivariant convolutions in the spherical domain. However, projection of data to a single sphere may be insufficient and lossy for more complex 3D data, e.g. when describing highly non-convex objects.
We also argue that representation by a single sphere is unnecessarily restrictive. We translate the desiderata to propose a new 3D representation architecture as follows: 
\begin{itemize}[nolistsep,nosep]
\item
Concentric spherical discretization, made up of multiple spheres at different radii, to more natively capture data according to their distribution in 3D space
\item
Efficient convolutions for learning to incorporate information over concentric spheres in a rotationally equivariant manner
\item
Hierarchical architecture to learn a representation over multiple scales, analogous to CNNs for 2D data
\end{itemize}

\begin{figure*}[t]
\center 
\includegraphics[width=1.0\textwidth]{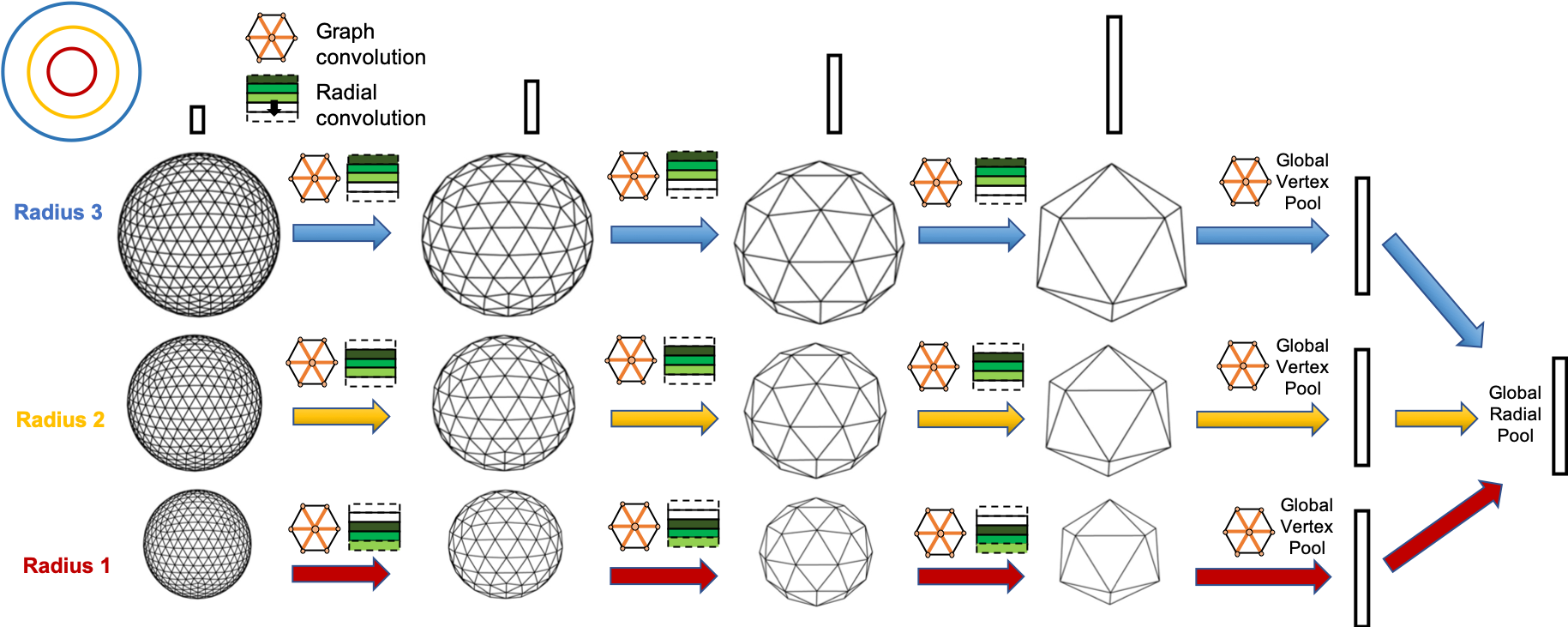}
\vspace{-2mm}
\caption{Example multi-radius architecture with $R=3$ concentric spheres. Graph convolutions are followed by radial convolutions at each level of coarsening. Graph convolution is applied within each sphere. Radial convolution (in this example) looks at 3 consecutive radial levels, and padding is applied to maintain radial dimensions across convolutions. Vertex pooling (not shown) then coarsens the discretization. Global vertex and radial pooling are applied at the end to obtain a final feature representation.}
\label{fig:model-MR}
\vspace{-3mm}
\end{figure*}

\section{Architecture Design}
To address the limitations of previous architectures and to increase capacity to distinguish different 3D data distributions, we introduce a new discretization based on concentric spheres, which additionally discretizes 3D space in the radial dimension.
We use a icosahedral discretization for each sphere, as it is highly uniform, permits efficient convolutions, and has a natural coarsening hierarchy.
To combine information in the concentric discretization, we propose (1) intra-sphere convolutions and (2) inter-sphere convolutions, which idividually operate on spatially orthogonal information.
We use graph convolutions for learning intra-sphere information, and radial convolutions for learning inter-sphere information.
Vertex pooling is applied to learn a hierarchical representation, following the discretization hierarchy of the icosahedron itself.

Fig. \ref{fig:model-MR} provides an illustrative overview of the resulting end-to-end architecture.
We explain each component of our architecture in more detail in subsequent sections.

\subsection{Concentric Spherical Discretization and Point Cloud Mapping}
In this section we explain the details of discretizing space by concentric spheres, which serves as the input spatial dimensions to the proposed  architecture.
We then explain our method for mapping arbitrary point cloud data to this discretization.

{\bf Spherical Discretization.}
\begin{figure}
\center 
\includegraphics[width=0.6\columnwidth]{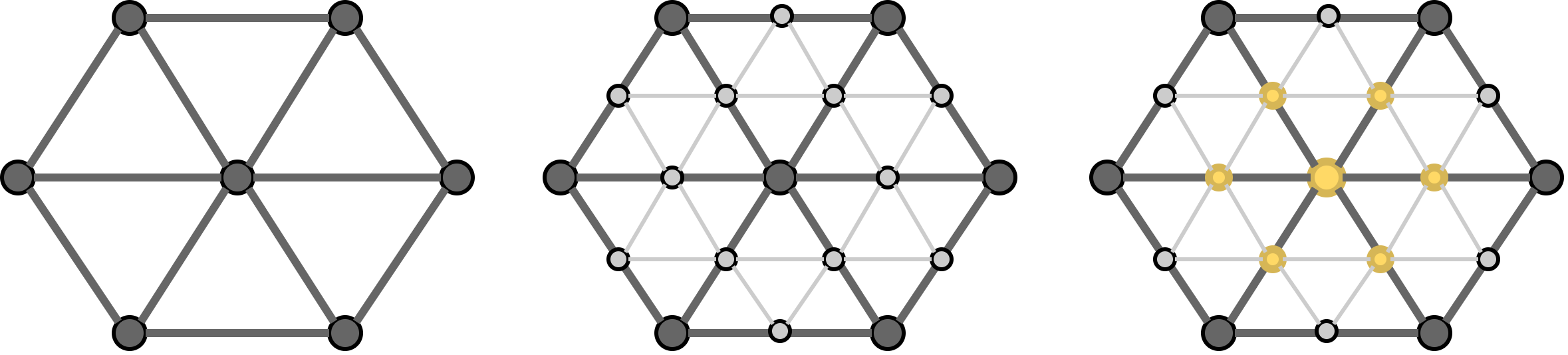}
\caption{The icosahedral grid (left) is defined by vertices spaced by equilateral triangles that can be recursively sub-divided to form a higher resolution grid (middle). This also defines a natural vertex and neighborhood hierarchy for pooling and coarsening in the reverse direction, where yellow highlighted vertices (right) are involved in pooling with the central vertex.}
\label{fig:pooling}
\end{figure}
\label{sec:icosahedron}
We work with an icosahedral grid discretization of the sphere. 
The base icosahedron $I^{(0)}$ has 12 vertices, forming 20 equilateral triangular face. 
To increase vertex resolution of the sphere, each face is evenly sub-divided into more equilateral triangles, with the number of vertices scaling as $|V| = 10*4^l+2$ (where $l$ is the target discretization level).
See Fig. \ref{fig:pooling} for illustration of the sub-division process.
Distances between adjacent vertices are no longer exactly equal once projected to the sphere, but overall the result is still a highly uniform spherical discretization.
The (single-channel) feature map corresponding to the discretization of a sphere can be defined by the vector $\vh \in \sR^{|V|}$.

\begin{figure}[h]
\center 
\includegraphics[width=0.75\columnwidth]{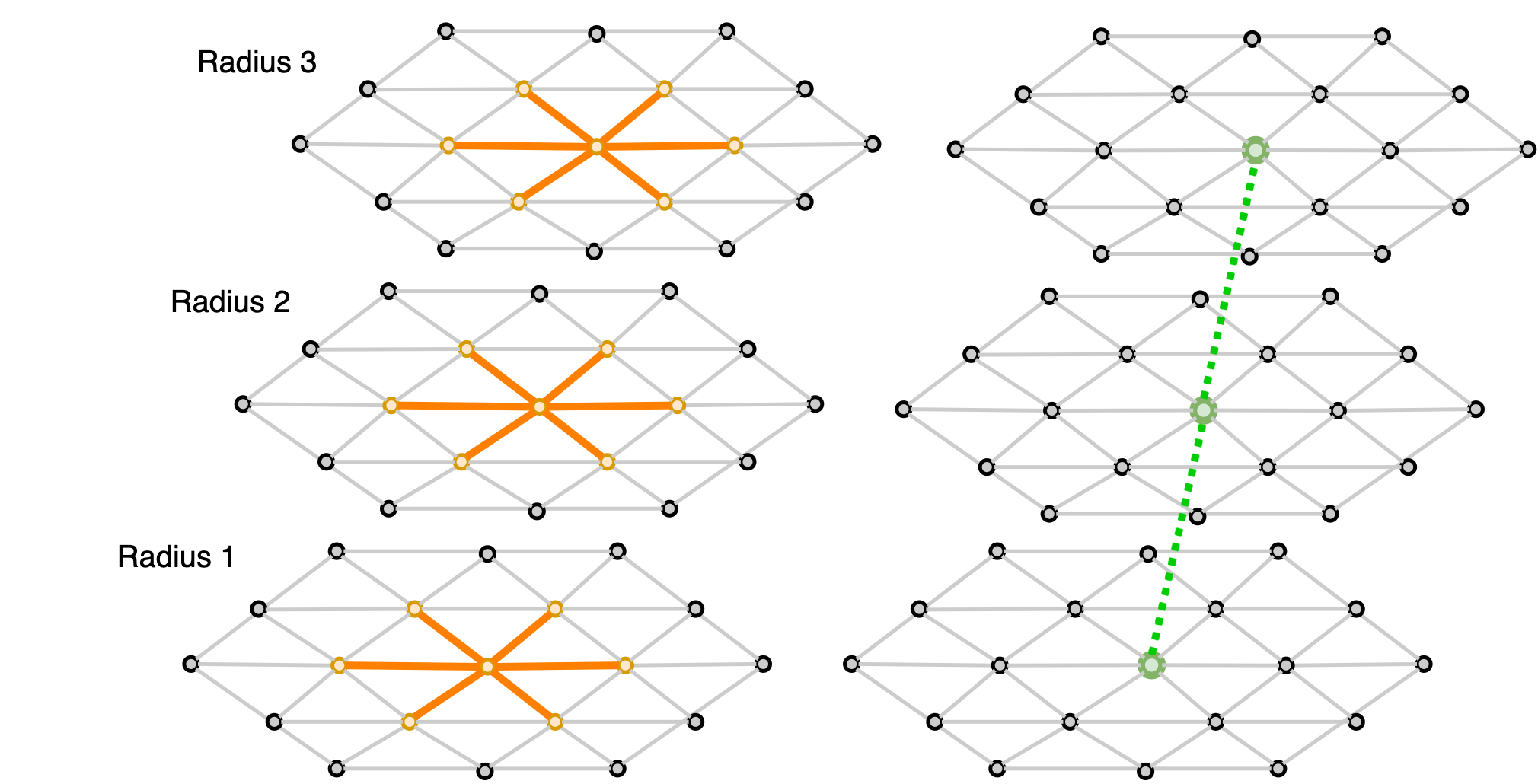}
\caption{Three spherical grids corresponding to different radius spheres are shown stacked on top of one another. Each vertex contains feature channels (not shown). Graph convolution (left) is applied with respect to vertices within the same sphere (example convolution neighborhood highlighted orange). Conversely, radial convolution is applied between vertices that are co-radial and between spheres (green highlight and dotted line).}
\label{fig:convolutions}
\end{figure}

{\bf Radial Discretization.}
We construct the multi-radius spherical discretization by stacking $R$ identical icosahedral grids along an additional radial dimension. 
See Fig. \ref{fig:convolutions} for illustration.
Each grid is an index of vertices belonging to a sphere of a particular radius in 3D space.
Furthermore, the same vertex on different grids are co-radial in 3D space.
Assuming unit radius normalization, we use a uniform discretization that results in concentric spheres scaled to radii $[\frac{1}{R}, \frac{2}{R}, ..., 1]$.
For a single-channel feature map, the result of stacking multiple spherical discretizations is a matrix $\mH \in \sR^{R\times|V|}$.

\begin{figure*}[t]
\center 
\includegraphics[width=1.0\textwidth]{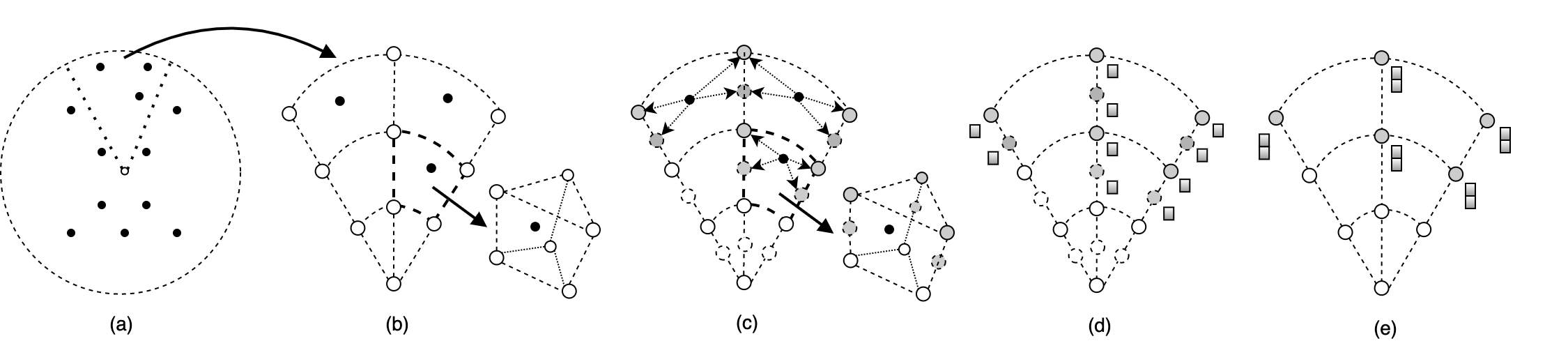}
\vspace{-2mm}
\caption{Illustrative example of mapping points to vertices. (a) shows an example point cloud (black dots) contained within a bounding sphere.
(b) shows the spherical partioning of 3D space in 2D cross section view, and zooms in on a sector occupied by 3 data points. Vertices are white circles. Each point is bounded by a neighborhood of 6 vertices, 3 from the sphere above and 3 from below.
(c),(d) Each point is mapped to scalar values defined on the bounding vertex neighborhood using radial basis functions.
Vertices affected by the mapping are shaded gray. Dotted circles indicate vertices temporarily added in the radial dimension to increase resolution.
(e) Vertex values are concatenated into feature channels of original vertices.}
\label{fig:mapping}
\vspace{-5mm}
\end{figure*}

{\bf Mapping Point Cloud to Concentric Spheres.}
With spherical and radial discretization in place, we now consider the problem of mapping a point cloud $\mP \in \mathbb{R}^{N \times 3}$ point cloud to an initial spherical feature map $\tH \in \mathbb{R}^{R \times |V| \times C}$, where $C$ is number of input channels.
While the concentric grid representation is defined discretely at fixed positions, the space of data point locations is continuous. We also aim to capture the distribution of points in a continuous way. 
To do so, we summarize the contribution of points using the Gaussian radial basis function (RBF):
\begin{align}
f(\vx) = \sum_{j=1}^N \phi(||\vx-\mP_j||_2^2)
\label{eq:rbf}
\end{align}
$N$ is the number of data points, and the function $\phi = \exp(-\gamma r^2)$ is parameterized by the bandwidth $\gamma$.
In practice we limit computation to a local neighborhood (instead of considering all points) and choose $\gamma$ accordingly. 
See Fig. \ref{fig:mapping} for visualization of the local neighborhood and mapping, and Sec. \ref{sec:mapping-appendix} for additional details on the choice of $\gamma$.

One possible mapping is to compute Eq. \ref{eq:rbf} at every vertex position of the spherical discretization, resulting in a single channel feature map. 
However, it is possible to obtain better resolution in capturing distribution of surrounding points by further sub-diving the discretized space, taking inspiration from \cite{meng2019vv}, along the radial dimension.
Subdividing the radial dimension by a factor $K_e$ results in a new spherical discretization with increased radial dimension of $R' = R*K_e$.
The RBF is evaluated at every vertex position of this new discretization, resulting in a feature map of dimension of $\tH' \in \mathbb{R}^{R' \times |V| \times 1}$. 
We map back to the original discretization by assigning $\tH_{i, u} = [\tH'_{j,u}: \ j\in(iK_e, iK_e+1, ..., 2iK_e, 2iK_e+1,..., 3iK_e-1)]$, resulting in a size  $[R \times |V| \times 2K_e]$ spherical feature map.
In summary, multiple RBF values are assigned to each vertex by further sub-dividing space in the radial dimension.

\subsection{Concentric Spherical Convolutions}
In this section we describe our implementation of intra-sphere and inter-sphere convolutions.
We also discuss intra-sphere vertex pooling and downsampling, which enables convolutions over different discretization scales in a hierarchical fashion.

{\bf Intra-sphere Convolutions.}
\label{sec:graph}
There is a growing body of work addressing design of rotation-equivariant filters over spherical feature maps.
We focus on graph convolutional filters for intra-sphere convolutions, as graph convolutions on icosahedral grids are scalable and rotationally equivariant up to discretization effects \parencite{cohen2019gauge,yang2020rotation}.
This motivates our construction of the undirected graph $G^{(l)} = (V^{(l)}, E^{(l)})$ from a level $l$ icosahedron $I^{(l)}$. Vertices of the vertex set $V^{(l)}$ correspond one-to-one with vertices of $I^{(l)}$ projected to unit sphere. $E^{(l)}$ is simply the set of all (bidirectional) face edges of the icosahedron (projected to unit sphere). Vertices are all degree 6, with exception of the the initial 12 vertices of the base icosahedron $I^{(0)}$ that are degree 5.
Since each edge is approximately equidistant between two points of the sphere \parencite{wang2011geometric}, $G^{(l)}$ is also treated as unweighted.
See Fig. \ref{fig:convolutions} for an example of neighbor connectivity.

With a well-defined connectivity graph and mapping to vertex features, we can now define graph convolution.
We use the mean-aggregator graph convolution operation defined in \textcite{hamilton2017inductive}, but our approach is not limited to the particular choice of graph convolution.
We introduce additional notation to define this convolution in the context of our  discretization in more detail.
Let $\tH \in \sR^{R \times |V| \times C}$ denote a $C$ channel tensor of features.
Also let $\mS, \mZ \in \sR^{C \times F}$ be shared parameters, where $C$ and $F$ are input and output channel sizes. $N(u)$ denote neighbors of vertex $u$ in graph $G$ and $\deg(u)$ denotes $|N(u)|$, the degree of vertex $u$.
We also introduce the subscript $t$ to indicate the convolutional layer number, $i \in [0, R-1]$ to index the radial dimension, and $u \in [0, |V|-1]$ to index the vertices.
The layer $t+1$ intra-sphere convolution output for vertex $u$ of sphere $i$ is then given by Eq. \ref{eq:intrasph}, where $\sigma$ indicates nonlinear activation function:
\begin{align}
\tH_{i,u}^{(t+1)} = \sigma\big (\tH_{i,u}^{(t)} \mS^{(t)} + (Mean_{v \in N(u)} \tH_{i,v}^{(t)}) \mZ^{(t)} \big)
\label{eq:intrasph}
\end{align}

{\bf Inter-sphere Convolutions.}
We introduce \emph{radial convolutions} to incorporate inter-sphere information, implemented as 1D convolutions where the radial dimension is treated as the sequence length.
Importantly, radial convolutions are rotationally invariant, as 1D convolution is operating over channels of the same vertex.
See Fig. \ref{fig:convolutions} for illustration. 
We introduce some additional notation to describe radial convolutions. Let $K$ be 1D convolution kernel size. We assume $K$ is odd valued, and pad inputs in the radial dimension such that a dimension of $R$ is maintained across convolutions.
Let $\mW \in \sR^{K \times C \times F}$ be a tensor of shared parameters, where $C$ and $F$ are input and output channel sizes.
The layer $t+1$ inter-sphere convolution output for vertex $u$ of sphere $i$ is given by Eq. \ref{eq:intersph}:
\begin{align}
\tH_{i,u}^{(t+1)} = \sigma(\sum_{k=-\lfloor\frac{K}{2}\rfloor}^{\lfloor\frac{K}{2}\rfloor} \tH_{i+k,u}^{(t)} \tW_{k+\lfloor\frac{K}{2}\rfloor}^{(t)})
\label{eq:intersph}
\end{align}

\subsection{Vertex Pooling}
Pooling is widely used alongside convolutional filters in CNN architectures to learn invariance to transformations of the input.
The icosahedron, due to its recursive refinement by discretization level, has a well-defined and natural hierarchy for pooling and coarsening.
Each coarser representation also follows a uniform discretization of the sphere, which allows efficient information propagation in hierarchical fashion when combined with convolutions and pooling.
Fig. \ref{fig:pooling} provides an example of the vertex pooling operation.
To formalize the pooling operation, we introduce the overloaded notation $\tH^{(l)} \in \sR^{R \times |V^{(l)}| \times C}$ to denote the feature tensor corresponding to $V^{(l)}$, the vertex set corresponding to level $l$ discretization.
Pooling is defined as $\tH_{i, u}^{(l-1)} = f(\{\tH_{i,v}^{(l)}: v \in N(u) \})$, where $N(u)$ is the neighborhood of vertex $u \in V^{(l)}$ and $f$ is a permutation invariant function (e.g. max operator).
Pooling is followed by downsampling, where only vertices of the smaller vertex set $V^{(l-1)}$ are retained.
We only apply neighborhood pooling within vertices of the same sphere.

\subsection{Complexity Analysis}
The neighborhood size of both graph and radial convolutions are constant, and so are their filter parameters due to weight sharing.
Therefore the overall complexity of both the graph convolution and radial convolution is $O(R\times|V|)$, or linear with respect to the total discretization size.
This introduces an additional factor $R$ of computational and memory cost compared to the $O(|V|)$ complexity of some spherical CNNs, corresponding to stacking multiple spherical discretizations.
However, as we show in experiments, $R$ is in practice a very small factor compared to the size of the spherical discretization.
It remains to be explored what performance tradeoff there is between $R$ and $|V|$ on a fixed discretization budget.

\section{Experiments}
\subsection{Point Cloud Classification}
\label{sec:modelnet}
\begin{figure}
\center 
\includegraphics[width=0.95\columnwidth]{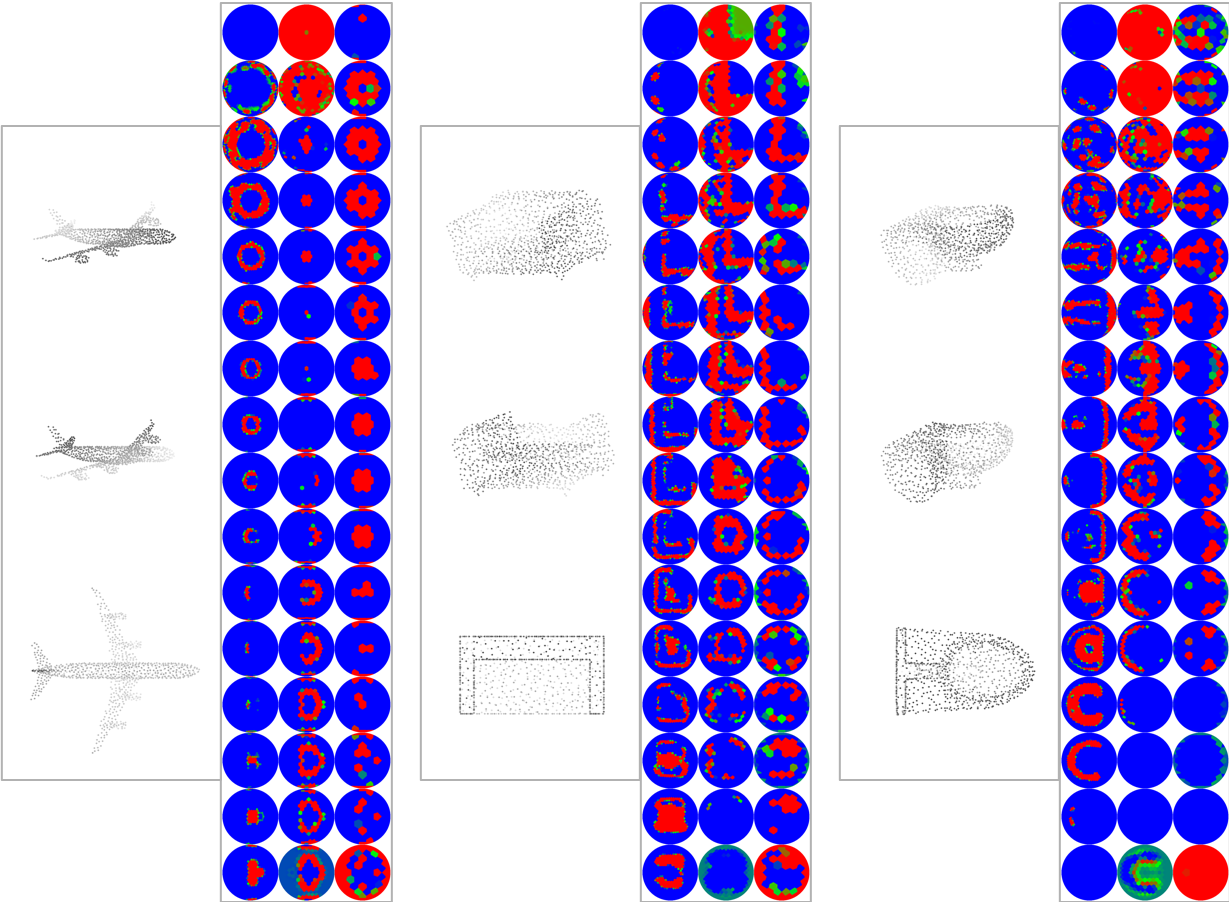}
\caption{Visualization of learned features of ModelNet40 point clouds. Example instances from left to right (each shown in 3 different orientations): airplane, sofa, and toilet. In the 16x3 tables, each sphere corresponds to a single feature channel. Rows correspond to radial level, with bottom rows corresponding to outer spheres. Columns correspond to discretization level of the sphere, from level 4 to 3 to 2 (left to right). Colors are interpolated between blue and red, corresponding to low or high normalized feature values. Different feature maps are captured at different radial levels; simultaneously there is also a high degree of continuity between consecutive spheres, suggesting there is information sharing between spheres resulting from radial convolutions.}
\label{fig:features}
\end{figure}

We consider the ModelNet40 3D shape classification task, with 12308 shapes and 40 classes. Each point cloud has 1024 points. For all experiments, 9840 shapes are used for training and 2468 for testing. See Fig. \ref{fig:features} for visualization examples of point clouds and our learned representation.

\begin{figure*}[h]
\center 
\includegraphics[width=1.0\textwidth]{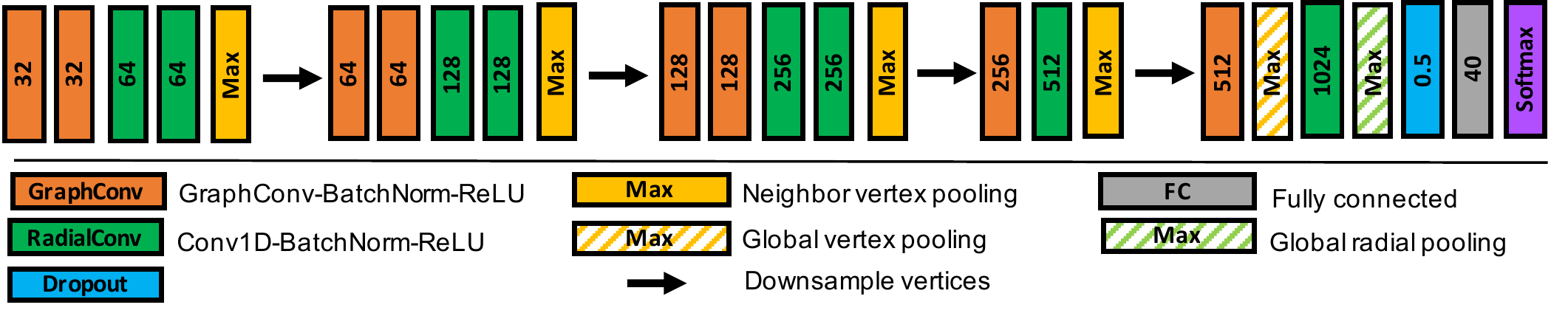}
\caption{Architecture for ModelNet40 classification. Initial feature channels is 16, resulting from point cloud RBF data mapping, defined over $L=4$ spherical discretization and $R=16$ spheres.
Number of output channels is shown where applicable.
1D convolution uses kernel size of 3. A residual connection is added between graph convolutions where the number of input and output channels match (not shown).}
\label{fig:modelnet-arch}
\end{figure*}

\tb{Architecture and Hyperparameters.}
Fig. \ref{fig:modelnet-arch} provides a detailed breakdown of architecture components for this experiment.
A level 4 icosahedron discretization and 16 concentric spheres are used to represent initial input.
Point clouds are mapped to vertex features using Gaussian RBF kernel with threshold $T=0.01$ and radial sub-division factor of $K_e=8$, resulting in 16 input channels.
The model is trained with Adam optimizer for 40 epochs using initial learning rate of 1e-3, along with learning rate decay by 0.1 after 25 and 35 epochs. Batch size is 32.

\begin{table*}[h!]
\center
\begin{tabular}{cccccc} \toprule
Method & Input & Params & z/z & SO3/SO3 & z/SO3 \\ \midrule
\emph{Pointwise Convolution} & & & & & \\
SRINet \parencite*{sun2019srinet} & xyz+ & 0.9M & 0.844 & 0.837 & 0.829 \\
& normal & & & & \\ 
PointNet \parencite*{qi2017pointnet} & xyz & 3.5M & 0.875 & 0.849 & 0.229\\
SPHNet \parencite*{poulenard2019effective} & xyz & 2.9M & 0.865 & 0.870 & 0.856  \\
RIConv \parencite*{zhang2019rotation} & xyz & 0.7M &  0.870 & 0.872 & \tb{0.870}\\ \midrule
\emph{Spherical CNN} & & & & & \\
PRIN \parencite*{you2018prin} & xyz &  1.7M & 0.819 & 0.810 & 0.765 \\ 
SFCNN \parencite*{rao2019spherical} & xyz & 9.2M & 0.888 & 0.874 & 0.831  \\ \midrule
\tb{Ours} (CSGNN) & xyz &  3.3M & \tb{0.896} & \tb{0.889} & 0.862  \\ \bottomrule
\end{tabular}
\caption{ModelNet40 point cloud classification results considering two different types of rotations: $z$-axis aligned, and more general $SO3$ rotations. For example, $SO3/SO3$ indicates training and testing with arbitrary rotations of input data. Number of parameters is in millions. CSGNN improves on the state of the art in both the $z/z$ setting and the more challenging and general $SO3/SO3$ setting.}
\label{tab:modelnet}
\end{table*}

{\bf Results.} We present our results and compare against related work in Table \ref{tab:modelnet} considering two types of rotations, following convention from earlier works: rotations about the $z$-axis, as well as as rotations with full degrees of freedom.
The latter is much more general and challenging, as it can encapsulate the full degree of objects in motion or data with no canonical orientation.
Our method outperforms others when training and testing in this setting ($SO3/SO3$) and achieve a 1.7\% relative improvement in accuracy over the next best result.
We also improve on state-of-the-art when training and testing on the less challenging rotations restricted to $z$-axis. 
Finally, our performance remains competitive with state of the art even when generalizing to unseen rotations in the $z/SO3$ setting.

We provide more detailed comparison with RIConv and SFCNN, which are best performing baselines from point-wise convolution and Spherical CNN categories.
RIConv uses rotationally invariant features at input and achieves very consistent performance regardless of the rotations seen or not seen.
While it achieves state of the art performance when generalizing to unseen rotations (z/SO3), its performance is less competitive when SO3 are accounted for in training. Our method outperforms it by a relative margin of 2.2\% in the $SO3/SO3$ setting.
Its performance falls even more behind when only testing with $z$-axis rotations, where our method records a 3\% relative improvement. 
This suggests some irrecoverable loss of expressiveness from the strict handling of rotational invariance at input.

SFCNN is most similar to our work in terms of also using graph-based convolutions over icosahedral spherical discretizaiton.
The key difference is that SFCNN is restricted to a single-sphere representation, instead relying on a PointNet-like module \parencite{qi2017pointnet} to learn a projection points to spherical features at input.
While this learned projection provides some descriptiveness, we achieve better results in all three settings by instead learning a concentric spherical representation.
While the Spherical CNN methods all suffer a performance drop when generalizing to unseen rotations, a limitation of the approach due to effects of discretization, SFCNN falls behind state of the art in the $z/SO3$ setting while our work remains competitive.

{\bf Evaluation details}:
For all experiments, a new rotation is sampled per instance in each epoch of training.
No other data augmentation is used to ensure cleaner comparison.
We report accuracy as average test scores across last 5 epochs of training to account for variation and the lack of a standard validation set.

\subsection{3D Mesh Classification}
\label{sec:shrec}
The SCHREC17 task has 51300 3D mesh models in 55 categories. We use the version where all models have been randomly perturbed by rotations.
\textcite{cohen2018spherical} presented a ray-casting scheme to regularly sample information incident to outermost mesh surfaces and obtain features maps defined over the spherical discretization. 
For sufficiently non-convex mesh objects, a single sphere projection may result in information loss, such as when a ray is incident to multiple surfaces occurring at different radii. 
This information is discarded by existing methods.
We propose a new data mapping that generalizes single sphere representation to a concentric spherical representation to preserve more information.

\label{sec:features}

\tb{Representation.} In the case of single-sphere representations, a single ray is projected from a source point (vertex) on the enclosing sphere towards the center of the object. 
The first hit incident with the mesh is recorded.
To extend ray-casting to multiple concentric spheres, we rescale the source point to the radii of each respective sphere. 
This results in multiple co-linear source points, one per sphere.
The 1st hit incident with the mesh is recorded for each ray cast from those source points, resulting in a multi-radius projection. 
While this new scheme is not sufficient to capture all incident surface information (e.g. if there are multiple hits sandwiched between two radial levels), it provides more samples that scales with the number of spheres.
We use a uniform $[\frac{1}{R}, \frac{2}{R}, ..., 1]$ radii division assuming inputs are normalized to unit radius. From each point of intersection with the mesh, the distance (with respect to outermost sphere) to the point of incidence as well as $sin$ and $cos$ features are recorded, resulting in 3 features per vertex.
\begin{table}[h!]
\center
\begin{tabular}{cccc} \toprule
Method & Params & $F_1$ micro & $F_1$ macro \\ \midrule
S2CNN \parencite*{cohen2018spherical} & 0.4M & 0.775 & 0.564 \\ \midrule
CSGNN ($R=1$) & 1.7M & 0.802 & 0.624  \\
CSGNN ($R=16$) & 3.7M & \tb{0.816} & \tb{0.638} \\ \bottomrule
\end{tabular}
\vspace{2mm}
\caption{SHREC17 classification performance in terms of $F_1$ micro-average and macro-average. CSGNN (this work) uses icosahedral spherical discretization with 2562 vertices. We our model with single sphere ($R=1$) and concentric spheres ($R=16$) with S2CNN (also single sphere).}
\label{tab:shrec}
\vspace{-4mm}
\end{table}

\tb{Architecture and Hyperparameters.}
The architecture for SHREC17 is identical to the one used for ModelNet40 in Fig. \ref{fig:modelnet-arch}, with the exception that the input dimension is 3 (corresponding to features obtained from ray-casting). 
We consider two model variations, single-sphere ($R=1$) and multi-sphere ($R=16$). 
In the single sphere version we use radial convolutional kernels of size 1, which is the same as applying connected layers with respect to each vertex's feature channels. We found that these additions to graph convolutions still boosted performance in the single-sphere case.
In the multi-sphere version we use a radial convolution kernel of size 3.
The former is trained for 20 epochs, with learning rate decay by factor of 0.1 after 15 and 20 epochs. 
The latter is trained for 25 epochs, with learning rate decay after 20 and 25 epochs.
Other training settings are same as for the ModelNet40 experiment.

\tb{Results.} 
See Table \ref{tab:shrec} for classification results.
S2CNN is a Spherical CNN for single sphere, and uses SO3 convolutions over a non-uniform discretization with 4096 grid points.
CSGNN uses graph convolutions over the highly uniform icosahedron discretization with 2562 vertices for intra-sphere convolutions.
We achieves significant performance improvement over S2CNN even when restricted to a single sphere in terms of both micro and macro $F_1$ score. 
This suggests that our use of graph convolutions combined with the icosahedral discretization already contributes a significant improvement.
We further improve performance by extending to 16 concentric spheres, obtaining 1.7\% and 2.2\% relative improvement in $F_1$ micro and macro scores over our single sphere version, respectively. 
Altogether, we obtain 5.3\% and 13.1\% relative improvement over S2CNN in the two classification metrics.

\tb{Evaluation details.} We record accuracy as average test scores across last 5 epochs of training. No data augmentation is applied in all experiments; data from both training and test sets have already been rotated.

\subsection{Model Ablation Study}
In this section we further analyze some key components to the performance of our model, such as the number of concentric spheres, parameters, and the importance of radial convolutions alongside graph convolutions.
All experiments are based on the ModelNet40 classificaiton task. 
Unless specified otherwise, we use a base model with $R=16$ and $L=4$, and keep the total number of parameters identical across all versions of the model.

\begin{figure}[h]
\center 
\includegraphics[width=0.6\columnwidth]{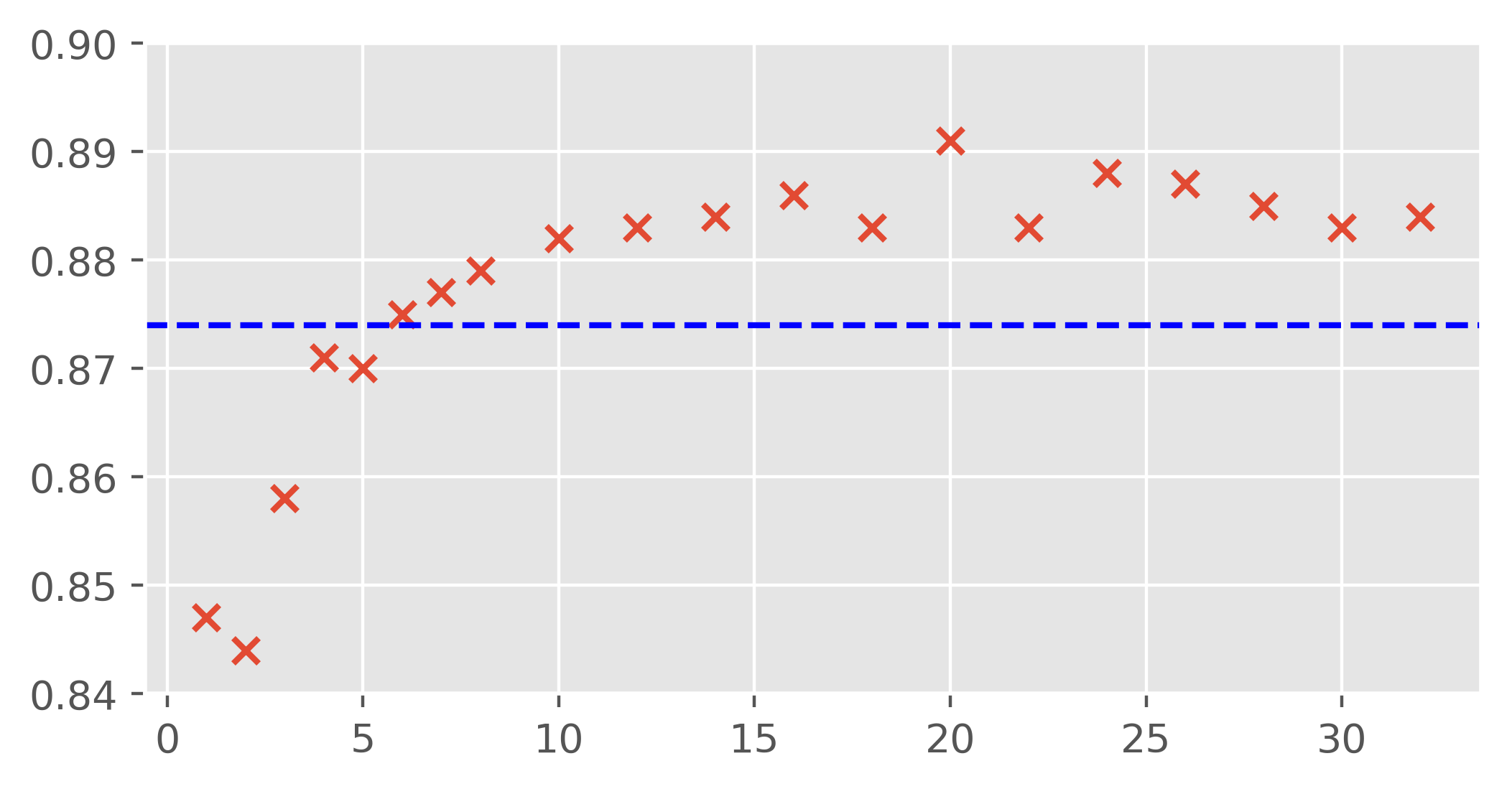}
\caption{Number of concentric spheres used (x-axis) vs. ModelNet40 point cloud classification accuracy on general rotations (SO3/SO3). Dotted blue line indicates previous state of the art performance. Number of parameters is fixed.}
\label{fig:rdims}
\end{figure}

{\bf Number of concentric spheres.}
To study the impact of multi-radius spherical discretization, we vary the number of spheres $R$ and present results in Fig. \ref{fig:rdims}.
To further isolate impact, we restrict the initial data mapping to a single channel feature and set $\gamma=0$ (otherwise some multi-radius information is still captured in initial input).
The architecture is the same as the one in Sec. \ref{fig:modelnet-arch}.
Performance consistently improves with higher radial dimension, peaking around $R=20$ with a 5.2\% relative accuracy improvement over $R=1$ version.
Using 6 concentric spheres already achieves state-of-the-art performance on this dataset.
The total number of parameters and layers is fixed across all settings, therefore performance is due to increased representational capacity from more radial resolution.

\begin{table}[t]
\center
\begin{tabular}{cc} \toprule
Setting & SO3/SO3 \\ \midrule
\emph{Graph convolution only} & \\
$R=1$, $M_{GC}=1$ & 0.862 \\ \midrule
\emph{Radial convolution only} & \\
$K_{RC} = 3$, $M_{RC}=1$ & 0.868 \\ \midrule
\emph{Radial kernel size} & \\
$K_{RC} = 1$ & 0.869 \\
$K_{RC} = 3$ & \tb{0.889} \\ \bottomrule
\end{tabular}
\vspace{2mm}
\caption{Ablation study on ModelNet40. $K_{RC}$ is size of radial convolutional kernel, $M_{GC}$ and $M_{RC}$ are number of graph and radial convolutional layers per block. $R=16$ and $L=4$, unless stated otherwise.}
\label{tab:ablation}
\end{table}

{\bf Parameters.}
\begin{figure}[h]
\center 
\includegraphics[width=0.6\columnwidth]{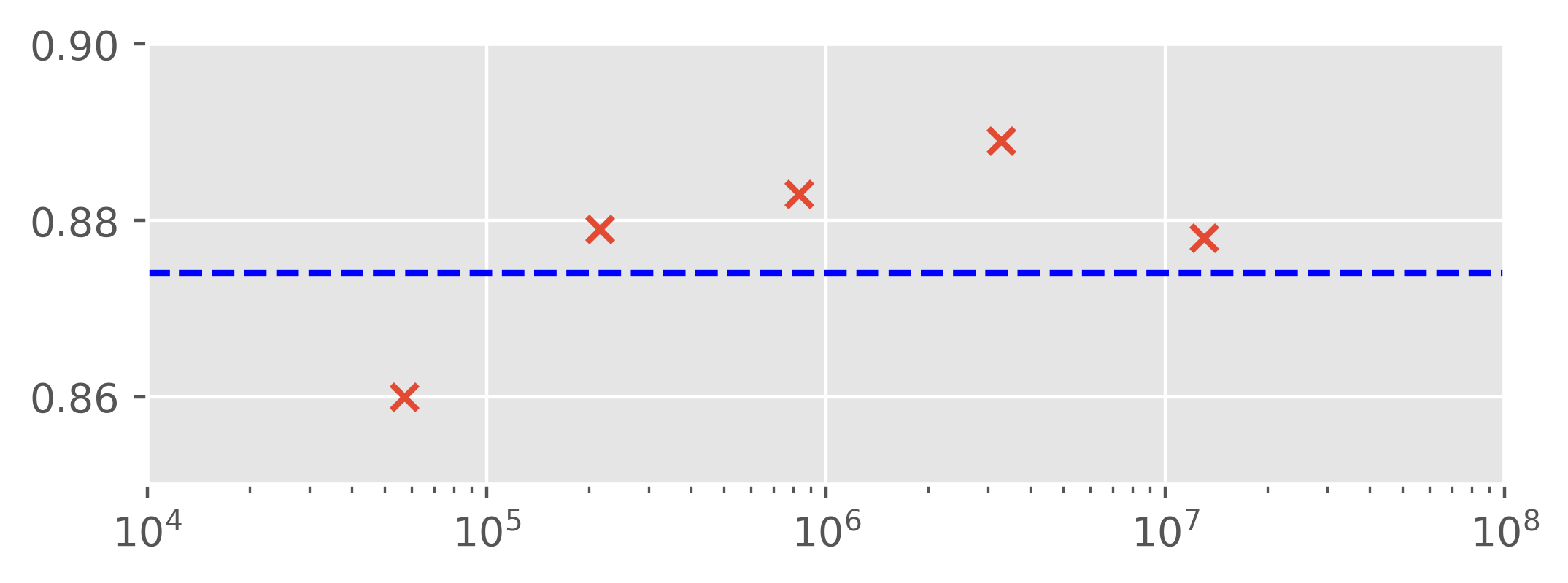}
\caption{Number of parameters (x-axis) vs. ModelNet40 point cloud classification accuracy for general rotations (SO3/SO3). Dotted blue line indicates previous state of the art performance. Number of spheres is fixed at $R=16$.}
\label{fig:params}
\vspace{-5mm}
\end{figure}
We vary the number of parameters used in our model (while keeping number of layers fixed) to evaluate impact on performance.
Our model already achieves state of the art performance on SO3/SO3 setting using only around 200K parameters (see Fig. \ref{fig:params}, which is less than the the number of parameters required from all other baselines from Table \ref{tab:modelnet}.

{\bf Radial convolutions.}
To evaluate the impact of radial convolutions, we first consider using only graph convolutions in a single-sphere discretization.
Even with capturing some radial information in the initial input using our data mapping, the result (see Table \ref{tab:ablation}) is considerably worse than that of our best version.
We also consider using only radial convolutions, which performs slightly better than using only graph convolutions over the single sphere.
These results suggest that it is essential to combine both types of convolutions for best performance.

Finally, we evaluate the importance of inter-sphere exchange of information via convolutions, compared to simply partitioning data in the radial dimension.
To do so, we compare using a radial convolutional kernel size of $K_{RC}=1$ vs. $K_{RC}=3$. $K_{RC}=1$ restricts the convolution to be within each radial level, while $K_{RC}$ widens the field of view to three consecutive levels.
Superior performance of $K_{RC}=3$ over $K_{RC}=1$ in Table \ref{tab:ablation} confirm that inter-sphere convolution is necessary for optimal performance.

\section{Discussion and Conclusions}
In this work we proposed a new multi-sphere convolutional architecture, CSGNN, for learning rotationally invariant representations of 3D data.
We introduced distinct intra-sphere and inter-sphere convolutions, which can be combined to learn more expressive representations compared to being restricted to single-sphere representation. 
Our use of graph and 1D convolutions preserves rotational equivariance, while achieving linear scalability with respect to size of discretization. 
We improve state-of-the-art performance in the most general rotation setting.
We also show that our approach generalizes to classification of 3D mesh objects by improving on single-sphere representation and performance for the SHREC17 task.
One avenue of future work is to explore more descriptive mappings of point cloud data to the discretization, as a learned assignment may better learn vertex features for describing nearby points. 
Exploring sparse convolutions by utilizing sparsity of point cloud data to improve scalability could be another avenue of future work.

\section{Acknowledgement}
Sandia National Laboratories is a multimission laboratory managed and operated by National Technology and Engineering Solutions of Sandia LLC, a wholly owned subsidiary of Honeywell International Inc. for the U.S. Department of Energy's National Nuclear Security Administration under contract DE-NA0003525.

\printbibliography

\appendix
\section{Appendix}
\subsection{Point Cloud to Spherical Signal}
\label{sec:mapping-appendix}
Instead of computing the summation in Eq. 1 with respect to all points, for each data point we update the features of vertices in a local neighborhood.
The radial basis function (RBF) $\phi$ decays exponentially, and so points beyond a local neighborhood have little influence (depending on choice of bandwidth $\gamma$).
Restricting to a constant size local neighborhood improves computation from $O(N|V|)$ to $O(N)$.

To define the local neighborhood of data point $p$ in this work: any given point $p$ is contained within two bounding ``triangles'' of the discretization (ignoring boundary conditions and degenerate cases). These correspond to the vertices $S^{(i)} = \{u^{(i)}, v^{(i)}, w^{(i)}\}$ and $S^{(i+1)}=\{u^{(i+1)}, v^{(i+1)}, w^{(i+1)}\}$, where $i$ indexes radial level.
However, using a single $\gamma$ value for the RBF results in scaling inconsistency: distances between vertices progressively shrink moving to inner spheres.
Based on the assumption that RBF values should be invariant to scale, a different $\gamma$ and corresponding RBF is defined with respect to radial level.
To define $\gamma_i$, we use the maximum pairwise distance $d_{max}^{(i)}$ between vertices in $\{S^{(i)}, S^{(i+1)}\}$. Specifically, we set $\gamma_i = \frac{-\log T}{{d_{max}^{(i)}}^2}$, where $T$ is a lower bound target RBF value. 
For example, $T=1$ would correspond to $\gamma_i=0$, or a RBF value of 1 at any distance.
$T \in (0, 1]$ is a tuning parameter that enables toggling the overall sensitivity of the RBF to distances.
Based on the approximation that $d_{max}^{(i)}$ is similar for any data point, $d_{max}^{(i)}$ is precomputed once.

\subsection{ModelNet40 Time Analysis}
\label{sec:times}
\begin{table*}[h!]
\center
\begin{tabular}{cccccccc} \toprule
Baseline & CSGNN & PointNet & RIConv & SPHNet & SRINet & SFCNN & PRIN \\ \midrule
Training (hrs) & 6.8 & 3.3 & 1.9 & 2.4 & 4.1 & 4.8 & 6.1 \\
Inference (s) & 0.16 & 0.016 & 0.062 & 0.099 & 0.097 & 0.069 & 0.004 \\ \bottomrule
\end{tabular}
\caption{Time comparisons of baselines for ModelNet40 dataset. Total training times are reported in hours. Inference time, in seconds, is for batch size of 32. CSGNN is our model.}
\label{tab:times}
\end{table*}
We compare total training time and batch inference time of baselines for ModelNet40 point cloud classification.
Results are reported in Table \ref{tab:times}. 
All baselines were run on the NVIDIA Tesla P100 GPU. 
Total training time includes data loading and transformation time. Inference time is based on batch size of 32, and computed from averaging 32 different batches. 
The inference times reported, unlike with training times, do not include data loading or transformation time.

\end{document}